\def\year{2019}\relax
\documentclass[letterpaper]{article} 
\usepackage{aaai19}  
\usepackage{times}  
\usepackage{helvet}  
\usepackage{courier}  
\usepackage{url}  
\usepackage{graphicx}  
\usepackage{amsmath}
\usepackage{amssymb}
\usepackage{booktabs}

\DeclareMathOperator*{\gru}{\operatorname{f_{GRU}}}
\DeclareMathOperator\softmax{\operatorname{softmax}}

\newcommand{\newcite}[1]
{\citeauthor{#1}~\shortcite{#1}}

\begin{document}
%
\title{Knowledge-Aware Conversational Semantic Parsing Over Web Tables}
\author{Yibo Sun$^\S$\thanks{\ \ Work is done during internship at Microsoft Research Asia.},\ \
     Duyu Tang$^\ddag$,
     Nan Duan$^\ddag$,
     Jingjing Xu$^{\dag*}$,
	\bf Xiaocheng Feng$^\S$, Bing Qin$^\S$\\
	$^\S$Harbin Institute of Technology, Harbin, China\\
	$^\ddag$Microsoft Research Asia, Beijing, China \\
    $^\dag$MOE Key Lab of Computational Linguistics, School of EECS, Peking University\\
	{\small \tt \{ybsun,xcfeng,qinb\}@ir.hit.edu.cn}\\
	{\small \tt \{dutang,nanduan\}@microsoft.com}\\
    {\small \tt \{jingjingxu\}@pku.edu.cn}\\
}
\maketitle
\begin{abstract}
Conversational semantic parsing over tables requires knowledge acquiring and reasoning abilities, which have not been well explored by current state-of-the-art approaches.
Motivated by this fact, we propose a knowledge-aware semantic parser to improve parsing performance by integrating various types of knowledge.
In this paper, we consider three types of knowledge, including grammar knowledge, expert knowledge, and external resource knowledge.
First, grammar knowledge empowers the model to effectively replicate previously generated logical form, which effectively handles the co-reference and ellipsis phenomena in conversation
Second, based on expert knowledge, we propose a decomposable model, which is more controllable compared with traditional end-to-end models that put all the burdens of learning on trial-and-error in an end-to-end way. Third, external resource knowledge, i.e., provided by a pre-trained language model or an entity typing model, is used to improve the representation of question and table for a better semantic understanding. We conduct experiments on the SequentialQA dataset.
Results show that our knowledge-aware model outperforms the state-of-the-art approaches. Incremental experimental results also prove the usefulness of various knowledge. Further analysis shows that our approach has the ability to derive the meaning representation of a context-dependent utterance by leveraging previously generated outcomes.
\end{abstract}

\section{Introduction}
We consider the problem of table-based conversational question answering, which is crucial for allowing users to interact with web tables or a relational databases using natural language.
Given a table, a question/utterance\footnote{In this work, we use the terms ``\textit{utterance}'' and ``\textit{question}'' interchangeably.} and the history of an interaction, the task calls for understanding the meanings of both current and historical utterances to produce the answer.
In this work, we tackle the problem in a semantic parsing paradigm \cite{Prolog1996LearningTP,wong2007learning,zettlemoyer2009learning,liang2016learning}.
User utterances are mapped to their formal meaning representations/logical forms (e.g. SQL queries), which could be regarded as programs that are executed on a table to yield the answer.
We use SequentialQA \cite{Iyyer2017SQA} as a testbed, and follow their experiment settings which learn from denotations (answers) without access to the logical forms.

\begin{figure}[t]
	\centering
	\includegraphics[width=.42\textwidth]{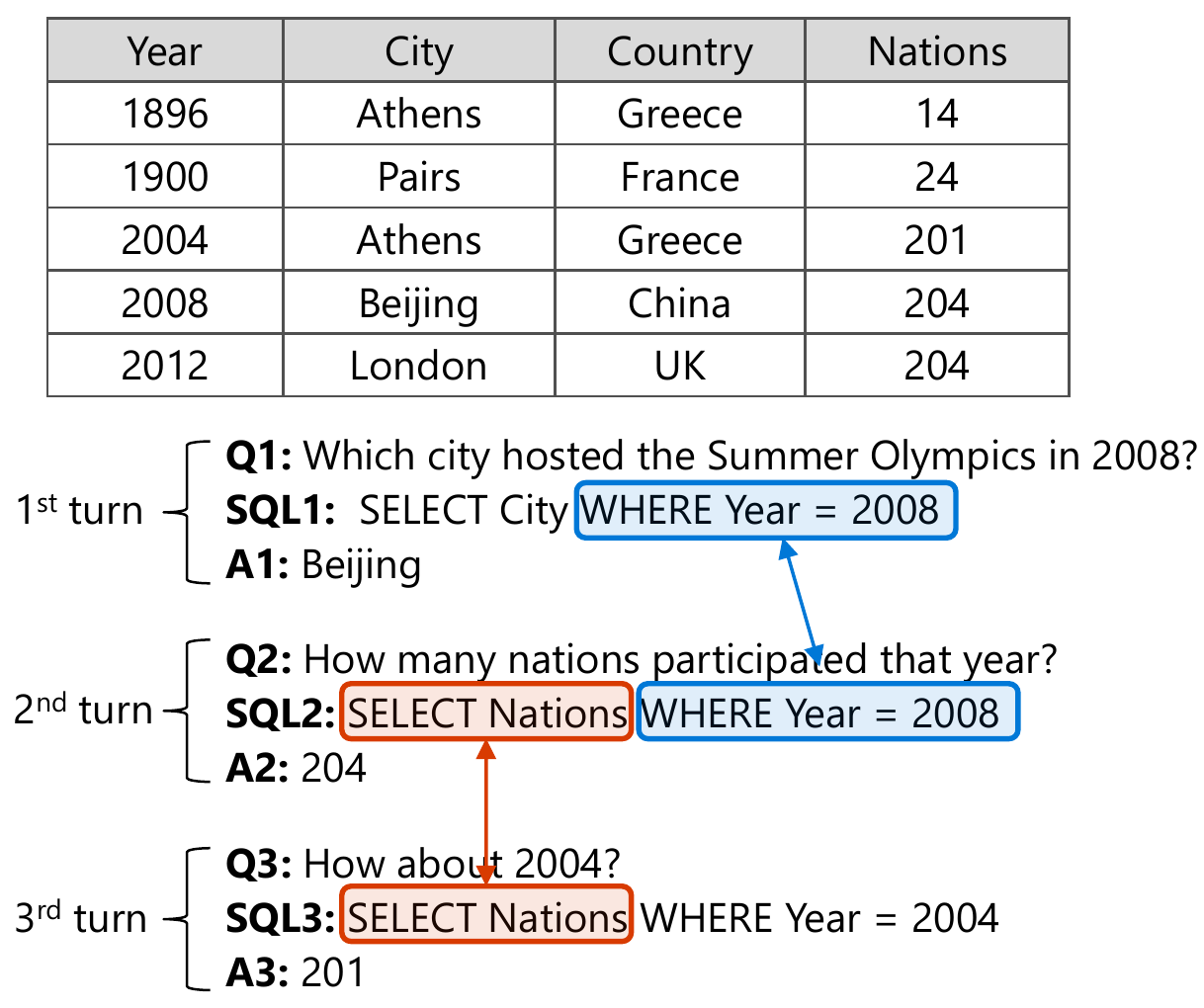}
	\caption{A running example that illustrates the input and the output of the problem.}
	\label{fig:intro-example}
\end{figure}

The task is challenging because successfully answering a question requires understanding the meanings of multiple inputs and reasoning based on that.
A model needs to understand the meaning of a question based on the meaning of a table as well as the understanding about historical questions.
Take the second turn question (``\textit{How many nations participate in that year?}'') in Figure \ref{fig:intro-example} as an example.
The model needs to understand that the question is asking about the number of nations with a constraint on a particular year.
Here the year (``\textit{2008}'') is not explicitly in Q2, but is a carry over from the analyzed result of the previous utterance.
There are different types of ellipsis and co-reference phenomena in user interactions.
The missing information in Q2 corresponds to the previous WHERE condition, while the missing part in Q3 comes from the previous SELECT clause.
Meanwhile, the whole process is also on the basis of understanding the meaning of a table including column names, cells, and the relationships between column names and cells.

Based on the aforementioned considerations, we present a \underline{c}onversational t\underline{a}ble-based se\underline{m}antic \underline{p}arser, abbreviated as CAMP, by introducing various types of knowledge in this work, including grammar knowledge, expert knowledge, and external resource knowledge.
First, we introduce grammar knowledge, which is the backbone of our model. Grammar knowledge includes a set of actions which could be easily used for reasoning and leveraging historical information.
We extend the grammar of \newcite{Iyyer2017SQA}, so that the model has the ability to copy logical form segment from previous outputs.
Therefore,
 our model effectively handles the co-reference and ellipsis phenomena in conversation, as shown in the second and third turns in Figure~\ref{fig:intro-example}.
Second, we use the expert knowledge to help us design model structure.
Considering that a decomposable model is more controllable, we decompose the entire pipeline into submodules which are coupled with the predefined actions in the grammar closely.
This further enables us to sample valid logical forms with improved heuristics, and learn submodules with fine-grained supervisions.
Third, we introduce several kinds of external resource knowledge to improve the understanding of input semantic meanings.
For a better question representation, we take advantage of a pre-trained language model by leveraging a large unstructured text corpus.
For a better table representation, we use several lexical analysis datasets and use the pre-trained models to give each table header semantic type information, i.e., NER type.

We  train model parameters from denotations without access to labeled logical forms, and conduct experiments on the SequentialQA dataset \cite{Iyyer2017SQA}.
Results show that our model achieves state-of-the-art accuracy.
Further analysis shows that (1) incrementally incorporating various types of knowledge could bring performance boost, and
(2) the model is capable of replicating previously generated logical forms to interpret the logical form of a conversational utterance.

\section{Grammar}
Our approach is based on a grammar consisting of predefined actions. We describe the grammar in this section and present our approach in the next section.
\begin{figure}[b]
	\centering
	\includegraphics[width=.46\textwidth]{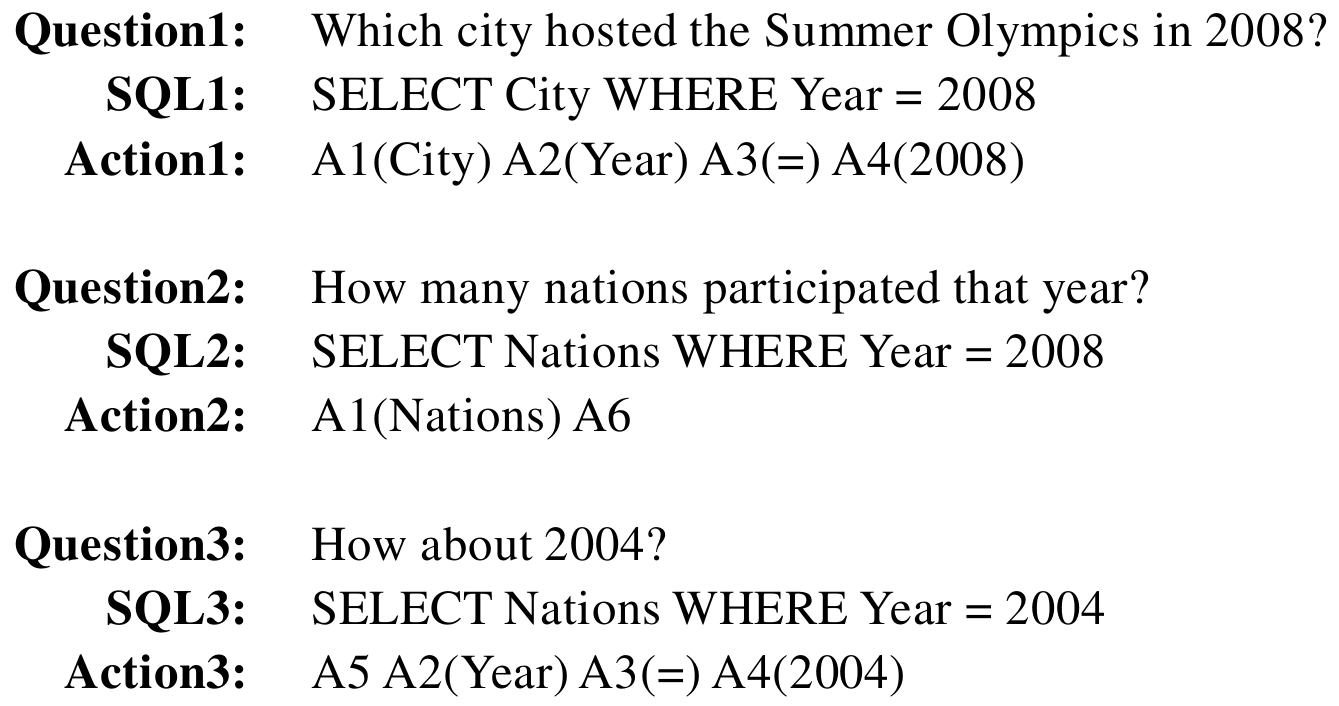}
	\caption{Examples of questions, corresponding SQL queries and actions. The bracket following an action represent its argument. }
	\label{fig:example}
\end{figure}

\begin{table}[t]
	\begin{tabular}{|c|l|l|}
		\hline
		Action & Operation     & \# Arguments \\ \hline
		A1     & SELECT-Col & \# columns          \\ \hline
		A2     & WHERE-Col  & \# columns          \\ \hline
		A3     & WHERE-Op  & \# operations          \\ \hline
		A4     & WHERE-Val  & \# valid cells          \\ \hline
		A5     & COPY SELECT  &   1          \\ \hline
		A6     & COPY WHERE  &   1          \\ \hline
		A7     & COPY SELECT + WHERE &  1            \\ \hline
	\end{tabular}
	\caption{Actions and the number of action instances in each type. Operations consist of
		$=, \neq, >, \geq, <, \leq, argmin, argmax$. A1 means selecting an column in SELECT expression. A2, A3 and A4 means selecting a column, a operation and a cell value in WHERE expression.
		A3 means select a condition operation, A4 means select a cell value, A5 means copying the previous SELECT expression, A6 means copying the previous WHERE expression, A7 means copying the entire previous SQL expression.}
	\label{tbl:action_types}
\end{table}
Partly inspired by the success of the sequence-to-action paradigm \cite{guu-EtAl:2017:Long,Bo2018SequencetoActionES,suhr2018learning} in semantic parsing, we treat the generation of a logical form as the generation of a linearized action sequence following a predefined grammar.
We use a SQL-like language as the logical form, which is a standard executable language on web tables.
Each query of this logical form language consist of one SELECT expression and zero or one WHERE expression. The SELECT expression shows which column can be chosen and the WHERE expression add a constraint on which row the answer can be chosen.
The SELECT expression consists of key word \textbf{SELECT} and a column name. The WHERE expression, which starts with the key word \textbf{WHERE}, consists of one or more condition expressions joined by the key word \textbf{AND}. Each condition expression consists of a column name, an operator, and a value. Following \newcite{Iyyer2017SQA}, we consider the following operators: $=, \neq, >, \geq, <, \leq, argmin, argmax$. Since the SequtialQA dataset only contains simple questions, we do not consider the key word AND, which means the WHERE expression only contains one condition expression.

\begin{figure}[h]
	\centering
	\includegraphics[width=.43\textwidth]{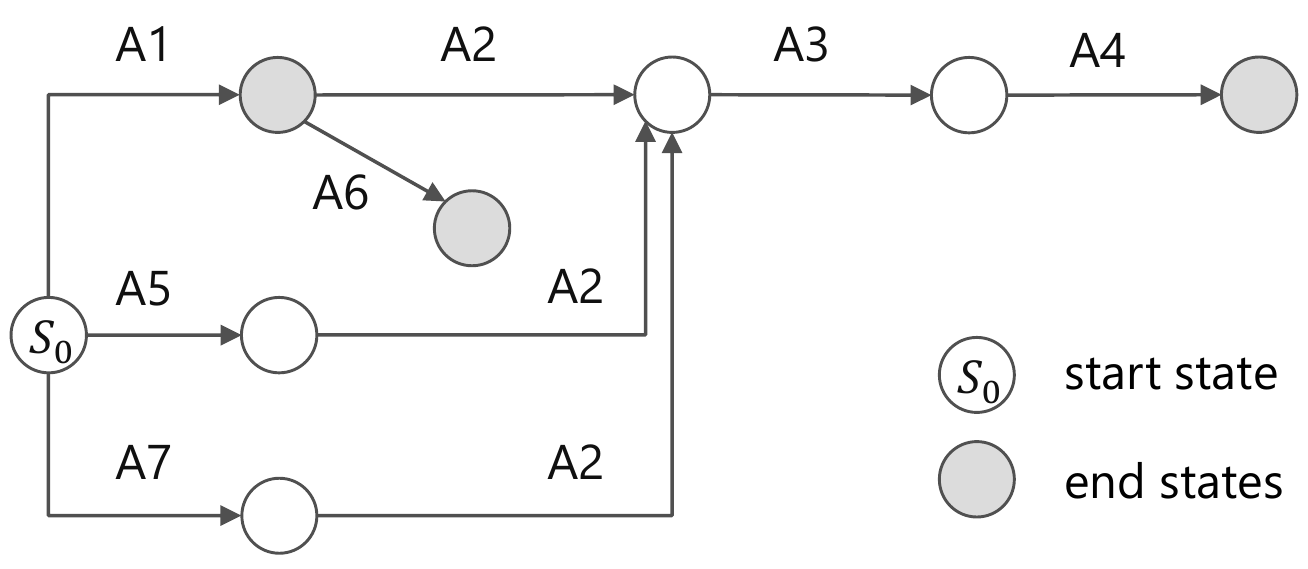}
	\caption{Possible action transitions based on our grammar as described in Table ~\ref{tbl:action_types}.}
	\label{fig:action-transitions}
\end{figure}

\begin{figure*}[t]
	\centering
	\includegraphics[width=.93\textwidth]{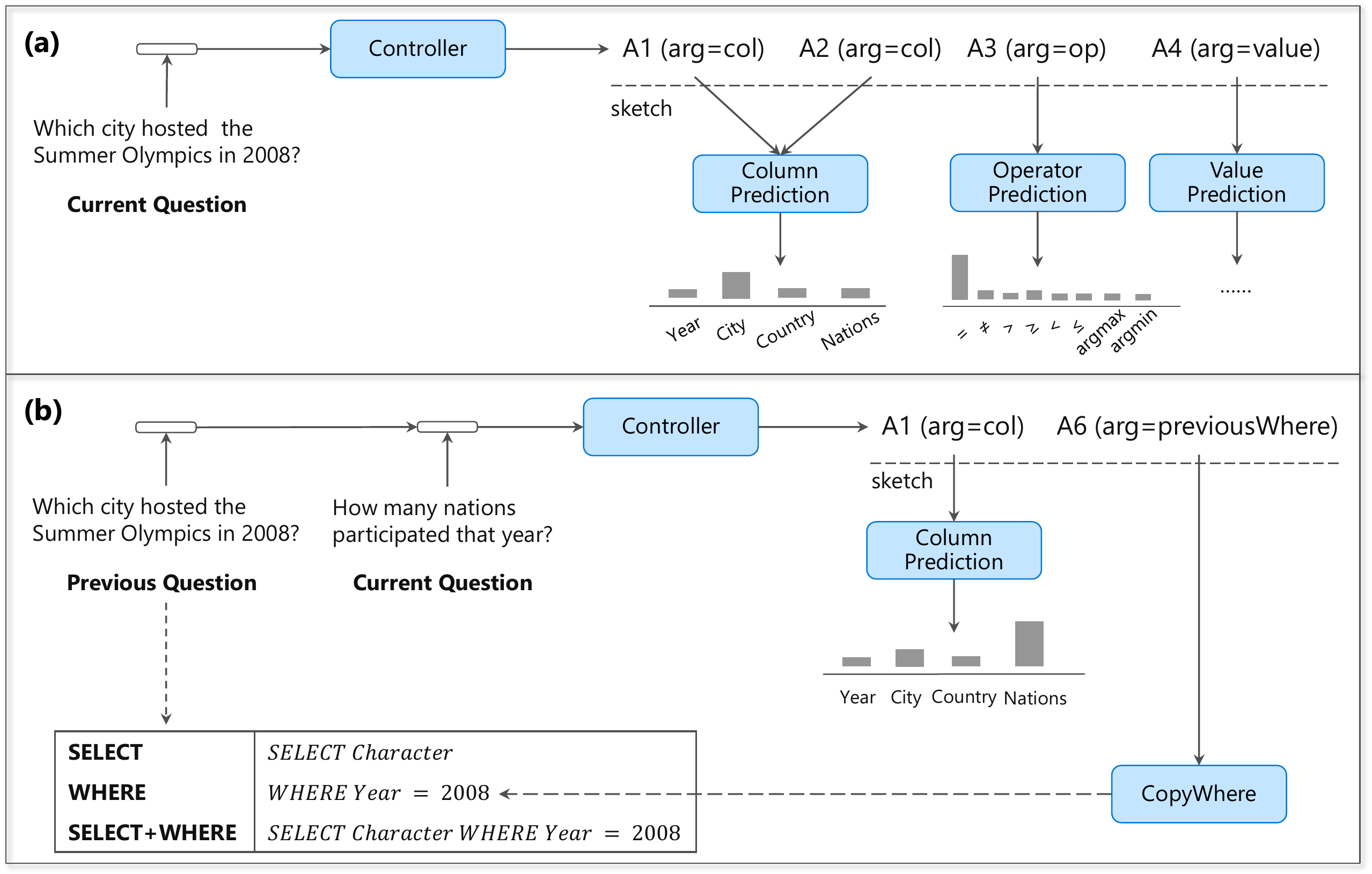}
	\caption{Overview of our model architecture.
		Part (a) shows how we handle a single turn question.
		Part (b) shows how we handle question that replicates logical form segment from the previous utterance.}
	\label{fig:model-example}
\end{figure*}
We describe the actions of our grammar in Table~\ref{tbl:action_types}. The first four actions (A1-A4) are designed to infer the logical forms based on the content of current utterance.
Thus they could be used to handle the context independent questions such as first question in Figure 2.
The last three actions are designed to replicate the previously generated logical forms.
For example, the second question in Figure 2 carry over the previous WHERE expression and interpret the SELECT expression based on the content of the second turn.
Similarly, the third question in Figure 2 use the the previous SELECT expression and infer the WHERE expression based on the third turn.
The numbers of arguments in copying actions are all equals to one because the SequtialQA dataset is composed of simple questions. Our approach can be easily extend to complex questions by representation previous logical form segment with embedding vector \cite{suhr2018learning}

\section{Approach}
We describe our \underline{c}onversational t\underline{a}ble-based se\underline{m}antic \underline{p}arser, dubbed as CAMP, in this section.

Given a question, a table, and previous questions in a conversation as the input, the model outputs a sequence of actions, which is equivalent to a logical form (i.e. SQL query in this work) which is executed on a table to obtain the answer.
Figure~\ref{fig:model-example} show the overview of our approach.
After encoding the questions into the vector representation, we first use a controller module to predict a sketch (which we will describe later) of the action sequence. Afterwards we use modules to predict the argument of each action in the sketch.
We will describe the controller and these modules in this section, and will also show how to incorporate knowledge these modules.

\paragraph{Question Encoder}
We describe how we encode the input question to a vector representation in this part.

The input includes the current utterance and the utterance in the previous turn.
We concatenate them with a special splitter and map each word to a continuous embedding vector
$\mathbf{x}_t^{I} = \mathbf{W}_x \mathbf{o}\left( x_t \right) $, where
$\mathbf{W}_x \in \mathbb{R}^{n \times |\mathcal{V}_x|}$ is an
embedding matrix, $|\mathcal{V}_x|$~is the vocabulary size, and
$\mathbf{o}\left( x_t \right)$~a one-hot vector.

We use a
bi-directional recurrent neural network with gated recurrent
units (\textsc{GRU}) ~\cite{Cho2014GRU} to represent the contextual information of each word. The
encoder computes the hidden vectors in a recursive way. The calulation of the $t$-th time
step is given as follows:
\begin{align}
\overrightarrow{\mathbf{e} }_{ t } &= \gru\left( \overrightarrow{\mathbf{e} }_{ t-1 } , \mathbf{x}_t \right) , t = 1, \cdots , |x| \label{eq:encoder:lstm:right} \\
\overleftarrow{\mathbf{e} }_{ t } &= \gru\left( \overleftarrow{\mathbf{e} }_{ t+1 } , \mathbf{x}_t \right) , t = |x|, \cdots , 1 \label{eq:encoder:lstm:left} \\
\mathbf{e}_t &= [\overrightarrow{\mathbf{e} }_{ t }, \overleftarrow{\mathbf{e} }_{ t }] \label{eq:encoder:lstm}
\end{align}
where $[\cdot,\cdot]$ denotes vector concatenation, $\mathbf{e}_t \in \mathbb{R}^{n}$, and $\gru$ is the GRU function.
We denote the input $x$ as:
$\tilde{\mathbf{e}} = [\overrightarrow{\tilde{\mathbf{e}}}_{ |x| }, \overleftarrow{\tilde{\mathbf{e}}}_{ 1 }]$

Inspired by the recently success of incorporating contextual knowledge in a variety of NLP tasks~\cite{Matthew2018elmo,Peters2018DissectLM},
we further enhance the contextual representation of each word by using a language model which is pretrained on a external text corpus.
In this work, we train a bidirectional language model from Paralex \cite{fader2013paraphrase}, which includes 18 million question-paraphrase pairs scraped from WikiAnswers.
The reason why we choose this dataset is that the question-style texts are more consistent with the genre of our input.
For each word $x$, we concat the word representaion and the hidden state $LM_t$ of the pretrained language model. $\mathbf{x}_t = [\mathbf{x}_t^{I}, LM_t]$

\paragraph{Table Encoder}
In this part, we describe how we encode headers and table cells into vector representations.

A table consists of $M$ headers (column names)\footnote{In this work, we use the terms ``\textit{header}'' and ``\textit{column name}'' interchangeably.} and $M * N$ cells where $N$ is the number of rows.
Each column consists of a header and several cells.
Let us denote the $k$-th headers as $c_k$.
Since a header may consist of multipule words, we use GRU RNN to calculate the presentation of each words and use
use the last hidden state as
the header vector representation  $\{\mathbf{c}_k\}_{k=1}^{M}$.
Cell values could be calculated in the same way.

We further improve header representation by considering typing information of cells.
The reason is that incorporating typing information would improve the predication of a header in SELECT and WHERE expressions.
Take Q2 in Figure~\ref{fig:intro-example} as an example. People can infer that the header ``Nation'' is talking about numbers because the cells in the same column are all numbers, and can use this information to better match to the question starting with ``how many''.
Therefore, the representation of a header not only depends on the words it contains, but also relates to the cells under the same column.
Specifically, we use  Stanford CoreNLP \cite{Manning2014TheSC} to get the NER result of each cell, and then use an off-the-shell mapping rule to get the type of each cell \cite{Li2002LearningQC}.
The header type is obtained by voting from the types of cells.
We follow \newcite{Li2002LearningQC} and set the types as \{COUNTRY, LOCATION, PERSON, DATE,
YEAR, TIME, TEXT, NUMBER, BOOLEAN, SEQUENCE, UNIT\}.

Formally, every header $\mathbf{c}_k$ also has a continuous type representation
via $\mathbf{t}_k = \mathbf{W}_t \mathbf{o}\left( c_t \right) $, where
$\mathbf{W}_t \in \mathbb{R}^{n \times |\mathcal{V}_t|}$ is an
embedding matrix, $|\mathcal{V}_t|$~is the total number of type, and
$\mathbf{o}\left( c_t \right)$~a one-hot vector.
The final representation of a header is the
concatenation of word-based vector and type-based vector $\mathbf{t_{k}}$, which we denote as follows.
\begin{equation}
\label{eq:wikisql:question}
\tilde{\mathbf{c_{k}}} = [\mathbf{c}_k, \mathbf{t}_k]
\end{equation}

\paragraph{Controller}
Given a current and previous question as input, the controller predict a sketch which is an action sequence without arguments between a starting state and an ending state.
As the number of all possible sketches we define is small, we model sketch generation as a classification problem.
Specifically, the sketch of [$A1$] means a logical form that only contains a SELECT expression, which is inferred based on the content of the current utterance.
The sketch of [$A1 \rightarrow  A2 \rightarrow A3 \rightarrow A4$] means a logical form having both SELECT expression and WHERE expression, in which case all the arguments are inferred based on the content of the current utterance.
The sketch of [$A5 \rightarrow A2 \rightarrow A3 \rightarrow A4$] stands for a logical form that replicates the SELECT expression of the previous utterance and infer out other constituents based on the current utterance.
The sketch of [$A6 \rightarrow A2 \rightarrow A3 \rightarrow A4$] represents a logical form that replicates previous the WHERE expression, and get the SELECT expression based on the current utterance.
The sketch of [$A7 \rightarrow A2 \rightarrow A3 \rightarrow A4$] means a logical form that replicates both SELECT expression and WHERE expression of the previous utterance, and incorporate additional constraint as another WHERE expression based on the current utterance.
Similar strategy has been proven effective in single-turn sequence-to-SQL generation \cite{Dong2018CoarsetoFineDF}.

Formally, given the current question $x_{cur}$ and the previous question $x_{pre}$, we use Equations ~\ref{eq:encoder:lstm} to get their vector representation $\tilde{\mathbf{e}}_{cur}$ and $\tilde{\mathbf{e}}_{pre}$.
We treat each sketch~$s$ as a category, and use a softmax classifier to
compute~$p\left( s | x \right)$ as follows, where $\mathbf{W}_a \in \mathbb{R}^{|\mathcal{V}_s| \times 2n} ,
\mathbf{b}_a \in \mathbb{R}^{|\mathcal{V}_s|}$ are parameters.
\begin{align}
p\left( s | x \right) = \softmax_{s} \left( \mathbf{W}_s [\tilde{\mathbf{e}}_{cur}, \tilde{\mathbf{e}}_{pre}] + \mathbf{b}_s \right) \nonumber
\end{align}

\paragraph{Column Prediction}
For action A1 and A2, we build two same neural models with different parameters. Both of them are used for predicting a column, just one in SELECT expressions and one in WHERE expressions.
We encode the input question~$x$
into $\{\mathbf{e}_t\}_{t=1}^{|x|}$ using GRU units.
For each column, we employ the column attention mechnism \cite{Xu2017SQLNetGS} to capture most relevant information from question. The column-aware question information is useful for column prediction.
Specifically, we use an attention mechanism towards question vectors
$\{\mathbf{e}_t\}_{t=1}^{|x|}$ to obtain the column-specific representation
for~$\mathbf{c}_k$. The attention score from $\mathbf{c}_k$ to
${\mathbf{e}}_{ t }$ is computed via ${ u }_{ k,t } \propto { \exp \{
  \alpha({\mathbf{c}}_{ k }) \cdot \alpha({\mathbf{e}}_{ t }) \} }$,
where $\alpha(\cdot)$ is a one-layer neural network, and $\sum _{ t=1
}^{ M }{ { u }_{ k,t } } = 1$. Then we compute the context vector
$\mathbf{e}_{k}^{c} = \sum_{ t=1 }^{ M }{ { u }_{ k,t } {\mathbf{e}}_{t } }$
to summarize the relevant question words for $\mathbf{c}_k$.

We calculate the probability of each column $\mathbf{c}_k$ via
\begin{equation}
\centering
\sigma(\mathbf{x}) = {\mathbf{w}_3 \cdot \tanh \left( \mathbf{W}_4 \mathbf{x} + \mathbf{b}_4 \right) } \label{eq:wikisql:agg_col:score}
\end{equation}
\begin{equation}
\centering
p\left( \texttt{col} =k | x \right) \propto  { \exp \{ \sigma([ [\tilde{\mathbf{e}},\mathbf{e}_{k}^{c} ] , {\mathbf{c}}_{ k } ]) \} } \label{eq:wikisql:agg_col}
\end{equation}
where $\sum _{ j=1 }^{ M }{ p\left( \texttt{col} = j | x \right) } = 1$, and $\mathbf{W}_4 \in \mathbb{R}^{3n \times m} , \mathbf{w}_3 , \mathbf{b}_4 \in \mathbb{R}^{m}$ are parameters.

\paragraph{Operator Prediction}
In this part, we need to predict an operator from the list [$=, \neq, >, \geq, <, \leq, argmin, argmax$].
We regard this task as a classification problem and use the same neural architecture in the controller module to make prediction.
For implementation, we randomly initialize the parameters and set the softmax's prediction category to the number of our operators, which is equals to 8 in this work.
\paragraph{Value Prediction}
We prediction WHERE value based on two evidences.
The first one comes from a neural network model which has the same  architecture  as  the one used for column prediction.
The second ones is calculated based on the number of word overlap between cell words and question words.
We incorporate the second score because we observe that many WHERE values are table cells that have string overlap with the question.
For example in Figure~\ref{fig:intro-example}, both the first and the third questions fall into this category.
Formerly, the final probability of a cell to be predicted is calculated as a linear combination of both distributions as following,
\begin{equation}
p\left(\texttt{cell} = k|x \right) = \lambda \hat{p}\left(\texttt{cell} = k|x \right) + (1-\lambda) \alpha^{cell}_k
\end{equation}
where $\hat{p}\left(\texttt{cell} = k|x \right)$ is the probability distribution obtained from the neural network and $\alpha^{cell}_k$ is the overlapping score  normalized by softmax and $\lambda$ is a hyper parameter.

\paragraph{COPYING Action Prediction}
As described in table~\ref{tbl:action_types}, we have tree copy-related actions (i.e. A5, A6, A7) to predict which component in the previous logical form should be copied to the current logical form. In this work this functionality is achieved by the controller model because the SequentialQA dataset only contains simple questions whose logical forms do not contain more than one WHERE expressions.
Our model easily extend to copy logical form segments from complex questions. An intuitive way to achieve this goal is representing each logical form component as a vector representation and applying an attention mechanism to choose the most relevant logical form segment \cite{suhr2018learning}.

\section{Training Data Collection}
We describe how we collect the supervisions from question-denotation pairs, which will be used to train each submodules in this section.

The SequentialQA dataset only provides question-denotation pairs, while our model requires question-action sequence pairs as the training data.
Therefore, we use the following strategies to automatically generate the logical form for each question, which is equivalent to an action sequence.
For acquiring the logical forms which are not provided by the SequentialQA dataset, we traverse the valid logical form space using breadth-first search following the action transition graph as illustrated in Figure~\ref{fig:action-transitions}.
For the purpose of preventing combinatorial explosion in searching pace, we prune the search space by executing the partial semantic parse over the table to get answers during the search process.
In this way, a path could be filtered out if its answers have no overlap with the golden answers. The coverage of our label generation process can be seen in Table~\ref{tbl:coverage}.

\begin{table}[h]
\centering

\begin{tabular}{l c c c}
	\toprule
	\textbf{\textsc{All}} & \textbf{\textsc{Pos 1}} & \textbf{\textsc{Pos 2}} & \textbf{\textsc{Pos 3}} \\
       \midrule
	   83.4  & 85.1   & 81.9   & 82.0  \\
\bottomrule
\end{tabular}
\normalsize
\caption{Percentage of questions in the training set that could be found at least one correct logical form.}
\label{tbl:coverage}
\end{table}

We use two strategies to handle the problem of spurious logical forms and to favor the actions of replicating from previous logical form segments, respectively.
The first strategy (\textbf{S1}) is for pruning spurious logical forms.
Spurious logical forms could be executed to get the correct answer but do not reflect the semantic meaning in the question.
Take question ``Which city hosted Summer Olympic in 2008?'' in Figure~\ref{fig:example} for example, a spurious logical form would be ``{SELECT} City {WHERE} Country {=} China''.
Pruning logical forms is vital for improving model performance according to previous studies \cite{Pasupat2016DPD,Mudrakarta2018ItWT}.
We only keep those logic forms whose components in the where clause have word overlap with the words in questions. In the previous example, the signal that ``2008'' appears in the question makes us choose ``SELECT City WHERE Year = 2008'' as the correct logical form.
The second strategy (\textbf{S2}) is for encouraging the model to learn sequential aspects of the dataset. We only keep the logical form with the COPY action after pruning spurious logical forms. The coverage of sampled correct logical forms in tracing set of SequentialQA is shown in Table~\ref{tbl:coverage}. The average numbers of logical forms for each turn before and after using the two strategies are given in Table~\ref{tbl:avg_length}.

\begin{table}[h]
\centering

\begin{tabular}{l c c c}
	\toprule
	\textbf{\textsc{Setting}} & \textbf{\textsc{Pos 1}} & \textbf{\textsc{Pos 2}} & \textbf{\textsc{Pos 3}} \\
     \midrule
	   Basic  & 2.08   & 3.80   & 5.19  \\
       Basic + S1  & 1.96   & 3.56   & 4.92  \\
       Basic + S1 + S2  & 1.96   & 2.20   & 3.40  \\
\bottomrule
\end{tabular}
\normalsize
\caption{Average number of logical forms for each question at different turns.}
\label{tbl:avg_length}
\end{table}

\section{Experiment}
We conduct the experiments on the SequentialQA dataset which has 6,066 unique questions sequences containing 17,553 total question-answer pairs (2.9 questions per sequence). The dataset is divided into train and test in an 83\%/17\% split.
We optimize our model parameters using standard stochastic gradient descent. We represent each word using word embedding~\cite{pennington2014glove} and the mean of the sub-word embeddings of all the n-grams in the
word (Hashimoto et al., 2016). The dimension of the concatenated word embedding is 400. We clamp the embedding values to avoid over-fitting.
We set the dimension of hidden state as 200, the dimension of type representation as 5, set the batch size as 32, and the dropout rate as 0.3.  We initialize model parameters from a uniform distribution with fan-in and fan-out. We use Adam as our optimization method and set the learning as 0.001.

\begin{table}[t]
	\centering
	\small
	\begin{tabular}{p{2.4cm}ccccc}
		\toprule
		\textbf{Model} & \textbf{\textsc{All}} & \textbf{\textsc{Seq}} & \textbf{\textsc{Pos 1}} & \textbf{\textsc{Pos 2}} & \textbf{\textsc{Pos 3}} \\ \midrule
		FP  & 34.1  & 7.2   & 52.6   & 25.6 & \textbf{25.9}  \\
		NP & 39.4  & 10.8  & 58.9  & 35.9 & 24.6  \\
		DynSP   & 42.0 & 10.2  & 70.9  & 35.8 & 20.1  \\
		\midrule
		FP+           & 33.2  & 7.7 & 51.4  & 22.2 & 22.3    \\
		NP+           & 40.2  & 11.8 & 60.0  & 35.9 & 25.5    \\
		DynSP*           & 44.7 & 12.8  & 70.4    & 41.1 & 23.6  \\
		\midrule
		CAMP & 45.0 & 11.7 & 71.3 & 42.8 & 21.9  \\
		CAMP + TU & \textbf{45.5} & 12.7 & \textbf{71.1} & \textbf{43.2} & 22.5  \\
		CAMP + TU + LM & \textbf{45.5} & \textbf{13.2} & 70.3 & 42.6 & 24.8  \\
		
		\bottomrule
	\end{tabular}
	\normalsize
	\caption{Accuracies of all systems on SequentailQA; the models in the top section of the table treat questions independently, while those in the middle consider sequential context. Our method in the bottom section also consider sequential context and outperforms existing ones both in terms of overall accuracy as well as sequence accuracy}
	\label{tbl:results}
\end{table}

\subsection{Baseline Systems}
We describe our baseline systems as follows.

\paragraph{Floating Parser.} \newcite{Pasupat15Compositional} build a system which first generates logical forms using a floating parser (FP) and then ranks the generated logical forms with a feature-based model. FP is like traditional chart parser but designed with specific deduction rules to alleviate the dependency on a full-fledged lexicon.

\paragraph{Neural Programmer.} The neural programmer (NP) \cite{neelakantan2016learning} is an end-to-end neural network-based approach.
NP has a set of predefined operations, and  Instead of directly generating a logical form, this system outputs a program consists of a fixed length of operations on a table. The handcraft operations is selected via attention mechanism \cite{Cho14attention} and the history information is conveyed by an RNN.

\paragraph{DynSP.} The dynamic neural semantic parsing framework~\cite{Iyyer2017SQA} constructs the logical form by applying predefined actions. It learns the model from annotations in a trial-and-error paradigm. The model is trained with policy functions using an improved algorithm proposed by \newcite{Peng2017MaximumMR}. DynSP* stands for an improved version that better utilize contextual information.

The experiment results related to FP and NP are reported by ~\newcite{Iyyer2017SQA}. The details of how they adjust these two models to the SequentialQA dataset can be seen in their paper.
We implement there variants of our model for comparision: CAMP is our basic framework where no external knowledge are used. In CAMP + TU, we incorporate type knowledge in table understanding module.
In CAMP + TU + LM, we further use contextual knowledge in question representation module.
\subsection{Results and Analysis}
\begin{table}[h]
	\centering
	\small
	\begin{tabular}{l c c c}
		\toprule
		\textbf{\textsc{models}} & CAMP &  + TU & + TU + LM \\
		\midrule
		{\textsc{Controller}}  & 83.5 & 83.5 & 84.8 \\
		{\textsc{SELECT-col}}          & 82.5 & 83.4 & 83.7 \\
		{\textsc{WHERE-col}}          & 35.0 & 35.9 & 36.6 \\
		{\textsc{Operation}}          & 69.7 & 69.7 & 70.2 \\
		{\textsc{Value}}          & 21.2 & 21.2 & 21.5 \\
		\bottomrule
	\end{tabular}
	\normalsize
	\caption{Accuracy for each module in different settings. }
	\label{tbl:each_model}
\end{table}

\begin{figure}[t]
	\centering
	\includegraphics[width=.48\textwidth]{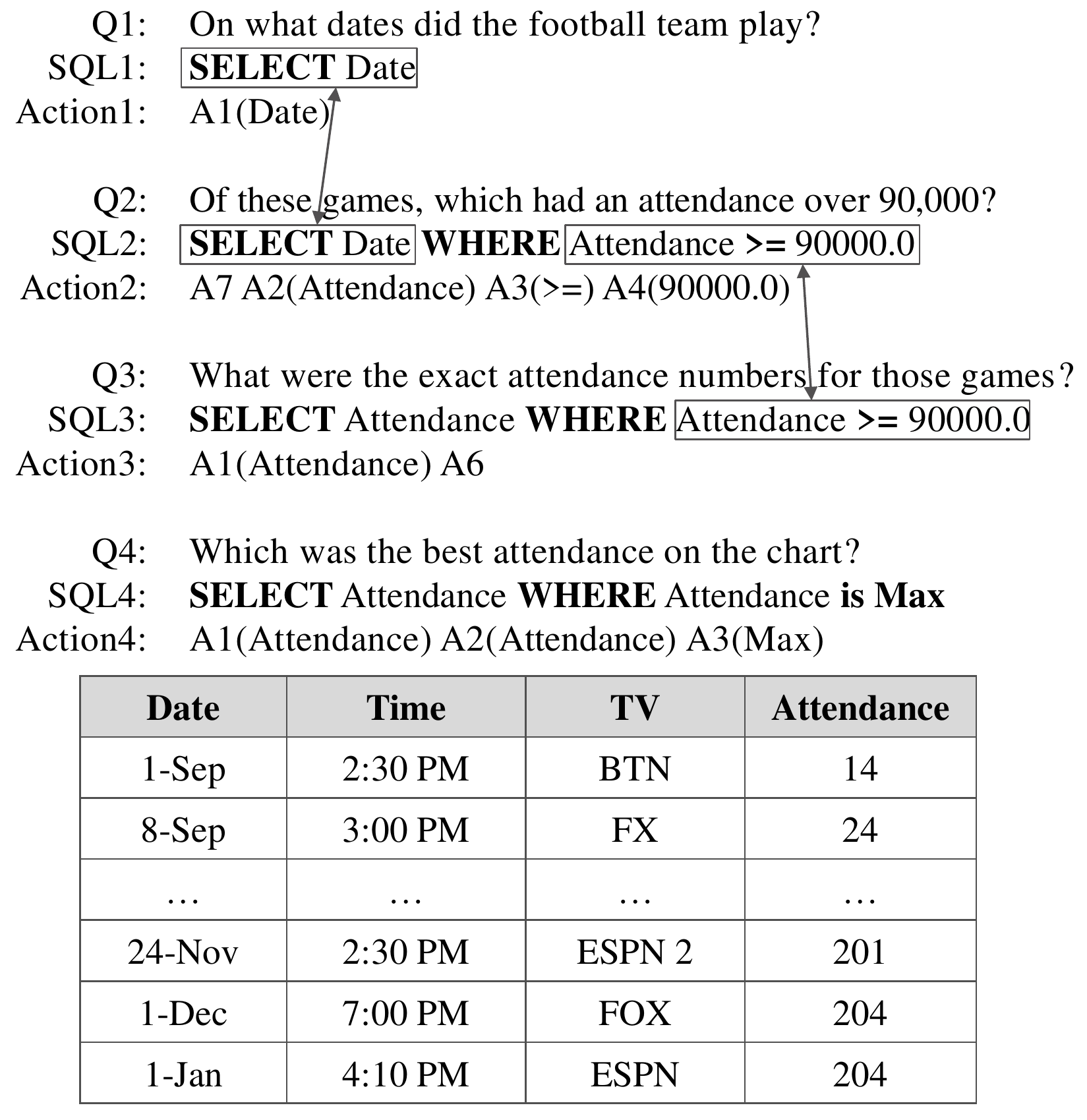}
	\caption{Outputs of a sequence of four sentences which are correctly produced by our approach.}
	\label{fig:correct_example}
\end{figure}

We can see the result of all baseline systems as well as our method in Table~\ref{tbl:results}. We show  accuracy for all questions, for each sequence (percentage of the correctly answered sequences of sentence), and for each sentence in particular position of a sequence.
We can see that CAMP performers better than existing systems in terms of overall accuracy. Adding table knowledge improves overall accuracy and sequence accuracy. Adding contextual knowledge further improve the sequence accuracy.
Figure~\ref{fig:correct_example} shows the outputs of a sequence of four sentences which are correctly produced by our approach.
We can see our model has the ability of effectively replicating previous logical form segment to derive the meaning representation of a context-dependent utterance.

We study the performance of each module of CAMP. From Table~\ref{tbl:each_model} we can see that table knowledge and contextual knowledge both bring improvements in these modules. The controller module and the column prediction module in SELECT expression achieves higher accuracies.
The reason is that, compared to other modules, the supervise signals for these two modules are less influenced by spurious logical forms.

\begin{figure}[h]
	\centering
	\includegraphics[width=.47\textwidth]{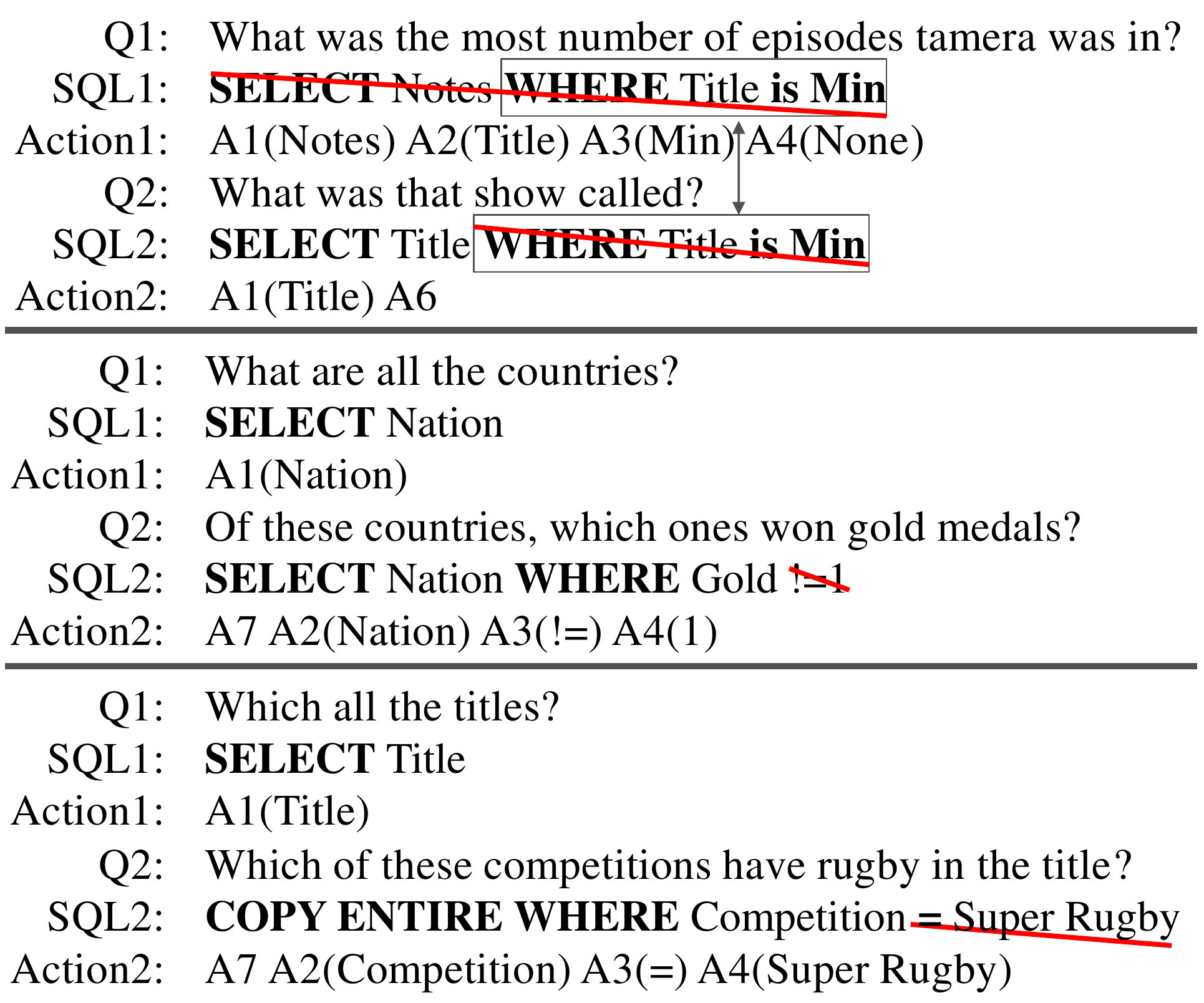}
	\caption{Wrongly predicated outputs by our model. Top: error progation;
	 Middle: SQL2 is wrong due to semantic matching error;
      Bottom: not coverd by current grammar.}
	\label{fig:error_example}
\end{figure}

We conduct error analysis to understand the limitation of our approach and shed light on future directions.
The errors are mainly caused by error propagation and semantic matching problems and limitation of our grammar.
Three examples are given in Figure~\ref{fig:error_example}.
In the top example the action sequence of question 2 is correctly predicted.
However the replicated WHERE clause form the previous logical form is incorrect.
In the middle example, ``won gold medals'' in the question should be interpreted as ``Gold $>$ 0''. However the term '``ones'' misleads the model to predict the WHERE value as ``1''.
The bottom example is unanswerable by our current model because our grammar as described in Table~\ref{tbl:action_types}. do not cover the ``contains in'' operator.

\section{Related Work}
This work closely relates to two lines of work, namely table-based semantic parsing and context-dependent semantic parsing. We describe the connections and the differences in this section.

Table-based semantic parsing aims to map an utterance to an executable logical form, which can be considered a program to execute on a table to yield the answer \cite{Pasupat15Compositional,Krishnamurthy2017NeuralSP,liang2018memory}.
The majority of existing studies focus on single-turn semantic parsing, in which case the meaning of the input utterance is independent of the historical interactions.
Existing studies on single-turn semantic parsing can be categorized based on the type of supervision used for model training.
The first category is the supervised setting, in which case the target logical forms are explicitly provided.
Supervised learning models including various sequence-to-sequence model architectures and slot filling based models have proven effective in learning the patterns involved in this type of parallel training data.
The second category is weak supervised learning, in which case the model can only access answers/denotations but does not have the annotated logical forms.
In this scenario, logical forms are typically regarded as the hidden variables/states.
Maximum marginal likelihood and reinforcement learning have proven effective in training the model \cite{guu-EtAl:2017:Long}.
Semi supervised learning is also investigated to further consider external unlabeled text corpora \cite{yin2018structvae}.
Different from the aforementioned studies, our work belongs to multi-turn table-based semantic parsing. The meaning of a question also depends on the conversation history.
The most relevant work is \newcite{Iyyer2017SQA}, the authors of which also develop the SequentialQA dataset. We have described the differences between our work and the work of \newcite{Iyyer2017SQA} in the introduction section.

In context-dependent semantic parsing, the understanding of an utterance also depends on some contexts.
We divide existing works based on the different types of ``context'', including the historical utterances and the state of the world which is the environment to execute the logical form on.
Our work belongs to the first group, namely historical questions as the context.
In this field, \newcite{zettlemoyer2009learning} learn the semantic parser from annotated lambda-calculus for ATIS flight planning interactions. They first carry out context-independent parsing with Combinatory Categorial Grammar (CCG), and then resolve all references and optionally perform an elaboration or deletion.
\newcite{vlachos2014new} deal with an interactive tourist information system, and use a set of classification modules to predict different arguments.
\newcite{suhr2018learning} also study on ATIS flight planning datasets, and introduce an improved sequence-to-sequence learning model to selectively replicate previous logical form segments.
In the second group, the world is regarded as the context and the state of the world is changeable as actions/logical forms are executed on the world.
\newcite{artzi2013weakly} focus on spatial instructions in the navigation environment and train a weighted CCG semantic parser.
\newcite{long-pasupat-liang:2016:P16-1} build three datasets including ALCHEMY, TANGRAMS and SCENE domains.
They take the starting state and the goal state of the entire instructions, and develop a shift-reduce parser based on a defined grammar for each domain.
\newcite{Suhr:18situated}
further introduce a learning algorithm to maximize the immediate expected rewards for all possible actions of each visited state.

\section{Conclusion}
In this work, we present a conversational table-based semantic parser called CAMP that integrates various types of knowledge.
Our approach integrates various types of knowledge, including unlabeled question utterances, typing information from external resource, and an improved grammar which is capable of replicating previously predicted action subsequence.
Each module in the entire pipeline can be conventionally improved.
We conduct experiments on the SequentialQA dataset, and train the model from question-denotation pairs.
Results show that incorporating knowledge improves the accuracy of our model, which achieves state-of-the-art accuracy on this dataset.
Further analysis shows that our approach has the ability to discovery and utilize previously generated logical forms to understand the meaning of the current utterance.

\bibliography{aaai2019}
\bibliographystyle{aaai}

\end{document}